\documentclass[lettersize,journal]{IEEEtran}
\usepackage{amsmath,amsfonts}
\usepackage{algorithmic}
\usepackage{algorithm}
\usepackage{array}
\usepackage{textcomp}
\usepackage{stfloats}
\usepackage{url}
\usepackage{verbatim}
\usepackage{graphicx}
\usepackage{subfigure}
\usepackage{float}
\usepackage{cite}
\usepackage{authblk}
\usepackage{color} 
\usepackage{hyperref}
\hypersetup{
     colorlinks = true,
     linkcolor = green,
     filecolor = green,
     citecolor = blue,      
     urlcolor = blue,
     }

\begin{document}

\title{KoMA: Knowledge-driven Multi-agent Framework for Autonomous Driving with Large Language Models}
\author[1]{Kemou Jiang}
\author[1]{Xuan Cai}
\author[1]{Zhiyong Cui*\thanks{* Corresponding Author}}
\author[1]{Aoyong Li}
\author[1]{Yilong Ren}
\author[1]{Haiyang Yu}
\author[2]{Hao Yang}
\author[3]{\\Daocheng Fu}
\author[3]{Licheng Wen}
\author[3]{Pinlong Cai*\thanks{}}
\affil[1]{State Key Laboratory of Intelligent Transportation Systems, School of Transportation Science and Engineering, Beihang University, 100191, Beijing, P.R.China.}
\affil[2]{Department of Civil and Systems Engineering, Johns Hopkins University, Baltimore, USA}
\affil[3]{Shanghai Artificial Intelligence Laboratory, 200232, Shanghai, P.R.China.}




\maketitle

\begin{abstract}
Large language models (LLMs) as autonomous agents offer a novel avenue for tackling real-world challenges through a knowledge-driven manner. These LLM-enhanced methodologies excel in generalization and interpretability. However, the complexity of driving tasks often necessitates the collaboration of multiple, heterogeneous agents, underscoring the need for such LLM-driven agents to engage in cooperative knowledge sharing and cognitive synergy. Despite the promise of LLMs, current applications predominantly center around single-agent scenarios, which limits their scope in the face of intricate, interconnected tasks. To broaden the horizons of knowledge-driven strategies and bolster the generalization capabilities of autonomous agents, we propose the $\mathsf{KoMA}$ framework consisting of the multi-agent interaction, the multi-step planning, the shared-memory, and the ranking-based reflection modules to enhance multi-agents' decision-making in complex driving scenarios. Based on the framework's generated text descriptions of driving scenarios, the multi-agent interaction module enables LLM agents to analyze and infer the intentions of surrounding vehicles based on scene information, akin to human cognition. The multi-step planning module enables LLM agents to analyze and obtain final action decisions layer by layer to ensure consistent goals for short-term action decisions. The shared memory module can accumulate collective experience to make superior decisions, and the ranking-based reflection module can evaluate and improve agent behavior with the aim of enhancing driving safety and efficiency. The $\mathsf{KoMA}$ framework not only enhances the robustness and adaptability of autonomous driving agents but also significantly elevates their generalization capabilities across diverse scenarios. Empirical results demonstrate the superiority of our approach over traditional methods, particularly in its ability to handle complex, unpredictable driving environments without extensive retraining. Project Page: \url{https://jkmhhh.github.io/KoMA/}.

\end{abstract}


\section{Introduction}
The quest for autonomous driving system has long been at the forefront of technological innovation, aiming to revolutionize transportation through improved safety, efficiency, and accessibility. Traditional approaches to autonomous driving have predominantly been data-driven\cite{amini2020learning,xu2020data,bogdoll2021description,chen2023milestones}, relying heavily on the collection and analysis of vast datasets to train algorithms capable of complex driving scenarios. While these methods have made significant strides, they are often hampered by challenges such as dataset bias\cite{jaipuria2020deflating}, overfitting\cite{ying2019overview,hawkins2004problem}, and a lack of interpretability\cite{zhang2021survey,gilpin2018explaining}, which can limit their effectiveness in novel or unforeseen circumstances.

In response to these challenges, there has been a paradigm shift towards knowledge-driven approaches in autonomous driving\cite{wendilu,li2023towards}. This shift is underpinned by the recognition that human drivers can rely on their rich experience and knowledge coupled with logical reasoning ability to make reasonable judgments and decisions when facing new scenarios. Large Language Models (LLMs) are trained on large amounts of text data to process, understand, and generate natural language text. The LLM has a broad range of basic human knowledge and superior reasoning abilities, making it a powerful tool in this new knowledge-driven paradigm\cite{touvron2023llama,achiam2023gpt}. These models, exemplified by the likes of GPT3.5 and GPT4\cite{achiam2023gpt} , have demonstrated unparalleled proficiency in understanding and generating natural language text and can quickly adapt to new application scenarios with a small number of prompt words, suggesting their potential to serve as agents within autonomous systems\cite{song2023llm,mai2023llm,pan2023context}. 

Recent studies on LLM-based autonomous driving agents have primarily been tested in simplistic scenarios, such as highway main road driving scenarios and circular track driving scenarios\cite{cui2024drive,wendilu,wang2023chatgpt}. In these contexts, the impact of other vehicles on the agent vehicle is negligible, placing the agent vehicle in a safe and stable environment. However, real-world driving scenarios are complex and time-variant, reflected by two major aspects: 1) the \textit{diversity of driving scenarios}, such as ramp merging and roundabouts, increases the likelihood of conflicts between vehicles due to their complexity. This diversity requires the vehicle agents to make reasonable and rapid plans to ensure driving safety and efficiency. 2) The second factor is the \textit{diversity of drivers} determined by the drivers' unique charateristics and reflected by their driving behaviors. This diversity contributes to the increased temporal variability of scenarios. Therefore, autonomous driving agents based on LLMs need further verification of their intelligence levels in complex and time-variant scenarios, especially when those agents with different objectives influence with each other. 

\begin{figure*}[!t]
    \centering
    \includegraphics[width=6in]{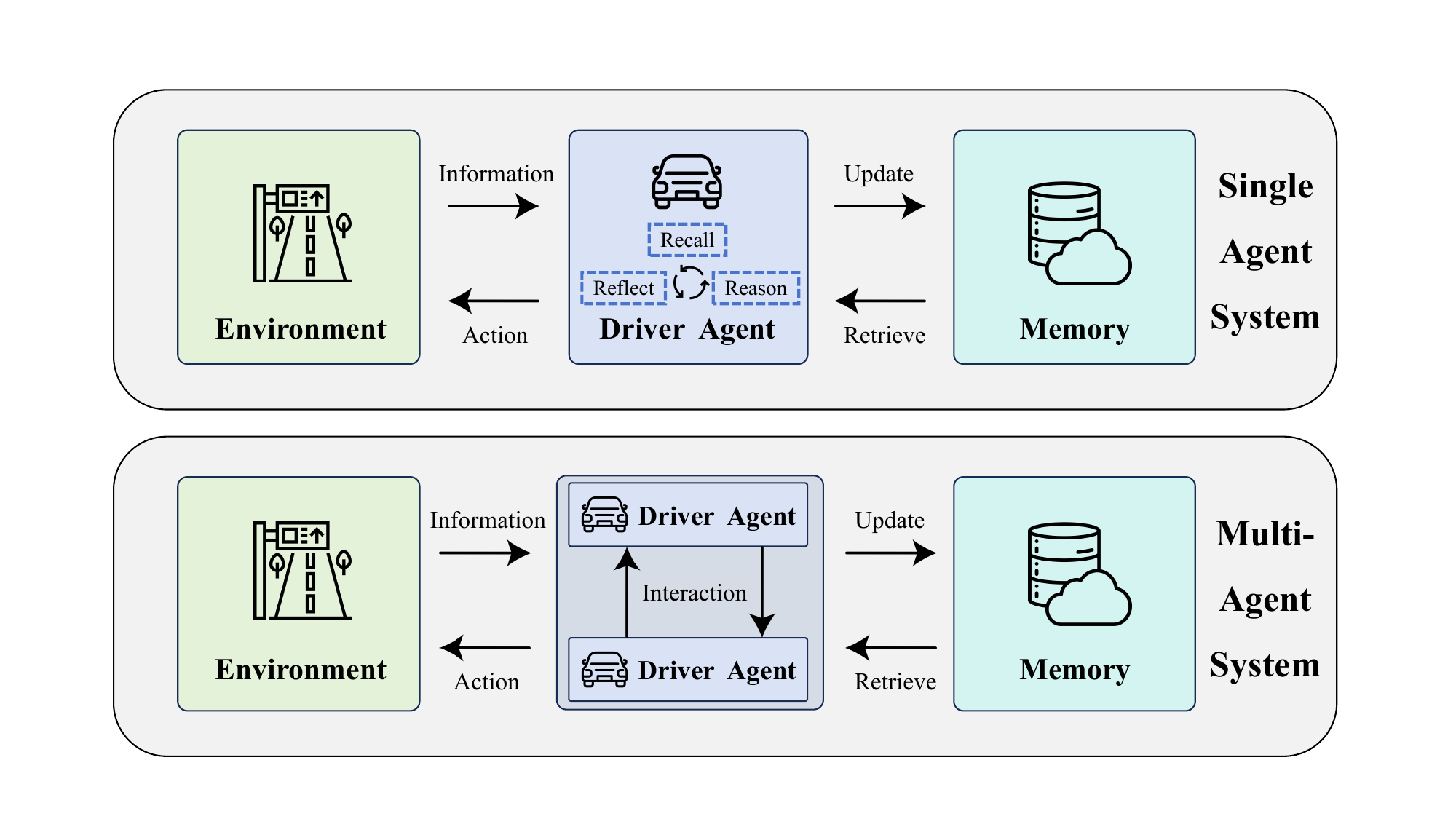}
    \caption{The knowledge-driven paradigm for single driving agent system and multiple driving agents system. Single knowledge-driven agent system including an interactive environment, a driver agent with recall, reasoning and reflection abilities, along with an independent memory module. Multiple knowledge-driven driving agent systems have an additional interaction module for communication and exchange among the agents.}
    \label{fig:0}
\end{figure*}

In multiagent driving scenarios, vehicle agents based on LLMs must account for both the impact of fixed driver model vehicles and the influence of variable LLM-based agent vehicles on their own driving decisions. By adopting this approach, the agents can more accurately simulate the complexities of real-world driving conditions, thereby advancing the frontier of knowledge-driven agent technology. The well-established knowledge-driven framework,  DiLu \cite{wendilu}, have summarized the knowledge-driven paradigm of autonomous driving systems, which comprises three components: 1) an environment with which the agent can interact; 2) a driving agent capable of memory, reasoning, and reflection; 3) a memory component to retain experience. As shown in Fig. \ref{fig:0}, the knowledge-driven paradigm can still be used within the multi-agent framework after expanding the interactions among multiple agents. However, due to the increased complexity of the scenarios in both time and space, further expansion and improvement of some modules are required to better apply them in complex scenarios. To address these limitations, this study introduce a knowledge-driven autonomous driving framework $\mathsf{KoMA}$ that incorporates multiple agents empowered by LLMs. It encapsulates five integral modules: Environment, Multi-agent Interaction, Multi-step Planning, Shared Memory, and Ranking-based Reflection.


At present, the reasoning module of LLMs is mainly divided into two approaches: making single-step decisions directly for each frame\cite{wendilu} and formulating a plan that includes a sequence of multiple-step decisions at once\cite{cui2024receive}. However, when faced with more complex driving scenarios, the former, which makes decisions for the next frame solely based on the current scene description, is prone to falling into local optimal solutions, lacking foresight, thereby making it difficult to achieve a goal that requires continuous action across multiple frames. In the face of more dynamic driving scenarios, the latter, which plans multiple-step action decisions at once, is susceptible to plan failure due to sudden changes in the scene conditions, thus failing to complete the scene goals smoothly. To address this, the $\mathsf{KoMA}$ framework introduces a multi-step reasoning module that achieves the final single-step action decision through a three-tiered reasoning process of goal-plan-action, ensuring the continuity of action decisions. 

In the knowledge-driven autonomous driving setting, the current reflection module is primarily activated after a collision occurs\cite{wendilu, li2023towards}, but this raises two issues: (1) In complex scenarios, the cause of an accident is often not the last action decision, but rather a critical decision made earlier in a long sequence of decisions. Identifying the key erroneous decision and reflecting on it for correction is a critical challenge. (2) In reality, drivers need to ensure not only the safety of driving but also its efficiency. An agent that completely sacrifices efficiency for safety is impractical for real-world scenarios. Therefore, the $\mathsf{KoMA}$ framework proposes a score-based reflection module that includes assessments of safety and efficiency. It expands the conditions for initiating reflection to situations where there is a sudden drop or extremely low score, allowing for timely correction of erroneous decisions and enhancing the quality of the memory fragments stored in the memory module.

The memory module serves as a repository for the historical driving experiences of an agent, capable of retrieving similar scenario experiences to assist the LLM-empowered agent in making action decisions. In early reinforcement learning multi-agent systems, each agent was trained independently\cite{tampuu2017multiagent}. However, this may lead to the collective intelligence being limited by an individual agent with poor training outcomes. Benefiting from the parameter sharing mechanism in reinforcement learning\cite{gupta2017cooperative}, the $\mathsf{KoMA}$ framework incorporates a shared memory module, allowing all agents to share a single vector database. This enables the sharing of driving experiences among agents, thereby enhancing the training speed and effectiveness of collective intelligence.

In summary, this study proposes $\mathsf{KoMA}$, a comprehensive framework that harnesses the power of LLMs to facilitate advanced decision-making in complex, multi-agent driving environments. The $\mathsf{KoMA}$ framework is designed to address the limitations of current single-agent approaches by integrating several key components that work in concert to enhance the capabilities of autonomous agents. Through a series of experiments and implementations, we demonstrate the feasibility of our approach and its advantages over traditional methods, particularly in terms of generalization, adaptability, and the ability to handle novel scenarios without extensive retraining.

The contribution of this paper is listed as follows:
\begin{enumerate}
    \item To the best of knowledge, we are the first to propose a knowledge-driven autonomous driving framework with multiple LLM-empowered agents, where those driving agents implicitly interact through estimating the intentions of surrounding vehicles. 
    \item We propose a three-layer structure of GPA (goal-planning-action) for analyzing tasks step by step in complex autonomous driving scenarios, ensuring the coherence of long-term decision-making. 
    \item The framework also integrating safety and efficiency indicators into the reflection module to accurately measure and locate erroneous decisions, thereby expanding the scope of reflection and enhancing the module's effectiveness.
    \item A memory sharing module is proposed to enable multiple agents quickly accumulate experience of different scenario at the same time and ensure that the memory and learning procedures of multiple agents are consistent. Experiments have demonstrated that this shared memory module effectively enhances the generalization capability of the agents.
\end{enumerate}

\section{Literature Review}

\subsection{LLM as Agent}
In the burgeoning field of artificial intelligence, the emergence and integration of LLMs as agents have marked a significant pivot from conventional rule-based systems to knowledge-driven approaches.
The LLMs demonstrate remarkable abilities to perform tasks based on user prompts and rapidly adapt to new scenarios through in-context learning. Zhou et al. highlighted the intrinsic properties of LLMs that facilitate this adaptability, laying the groundwork for their application as complex agents\cite{zhoulanguage}. The shift towarded employing LLMs as the ``brain" or controller of agents, as discussed by Xi et al., signified a pivotal transition towards more integrated and autonomous systems\cite{xi2023rise}.
Gao et al. introduced AssistGPT, which employed a method of linguistic reasoning called Plan, Do, Check, and Learn (PEIL). This methodology enhanced the integration of LLMs with various tools, pushing the boundaries of their application\cite{gao2023assistgpt}. Similarly, Shen et al. proposed HuggingGPT, a framework designed to leverage LLMs for connecting disparate AI models within the machine learning community to address complex AI tasks\cite{shen2024hugginggpt}. Furthermore, ViperGPT, as presented by Suris et al., showcased a novel approach by utilizing an API to access and compose modules through Python code generation, thereby offering solutions for an array of queries\cite{suris2023vipergpt}.
A significant evolution in the deployment of LLMs as agents was the transition from traditional, data-driven paradigms to knowledge-driven methodologies. Li et al. emphasized this shift, advocating for a move towards active, cognition-based understanding that leveraged the extensive general knowledge and reasoning capabilities of LLMs\cite{li2023towards}. This approach not only enhanced the agent's ability to interact with and comprehend the world, but also enabled the system to become more autonomous and intelligent with relevant domain knowledge and reasoning learning ability.
    
\subsection{Multi agent with LLM}

The integration of LLMs in multi-agent systems has become an emerging research field for improving collective intelligence and collaboration in various fields. The exploration of LLM-powered multi-agent systems revealed the future of collaborative intelligence, where the cooperation of multiple independent agents could enhance problem-solving capabilities.
Handler et al. proposed a multi-dimensional taxonomy to tackle the difficulties in categorize and understand the architectural complexities posed by LLM-powered multi-agent systems which aimed at accomplishing complex tasks, goals, or problems with the cognitive synergy of multiple autonomous LLM-powered agents\cite{handler2023balancing}.
Liu et al. proposed the Dynamic LLM-Agent Network (DyLAN), a novel approach for fostering LLM-agent collaboration on intricate tasks such as reasoning and code generation. DyLAN created a strategic assembly of agents that communicate within a dynamic interaction architecture tailored to the specific requirements of the task query\cite{liu2023dynamic}. This model exemplified the potential of context-aware collaboration among LLM-powered agents, pushing the boundaries of collective problem-solving capabilities.
Chen et al. introduced AGENTVERSE, a multi-agent framework designed to orchestrate a collaborative group of expert agents, creating a system whose capabilities exceed the sum of its parts. AGENTVERSE enhanced the efficiency and effectiveness of task accomplishment, showcasing the power of agent collaboration in complex problem-solving scenarios\cite{chen2023agentverse}.
Gong et al. presented Mindagent, an infrastructure that leverages LLMs for interactive multi-agent planning. This system not only demonstrated the in-context learning capabilities of LLMs in multi-agent planning but also offered several prompting techniques to bolster their planning proficiency. It is proposed for evaluating planning and coordination capabilities in the context of gaming interaction\cite{gong2024mindagent}.
Zhang et al. developed the Cooperative Embodied Language Agent (CoELA), an agent capable of planning, communicating, and cooperating with others to efficiently accomplish long-horizon tasks. Powered by GPT4, CoELA surpassed traditional planning-based methods, demonstrating emergent effective communication strategies. This advance highlighted the potential for LLM to conduct cooperative behavior and perform complex tasks in multiple agents\cite{zhangbuilding}.

\begin{figure*}[hb]
    \centering
    \includegraphics[width=7.5in]{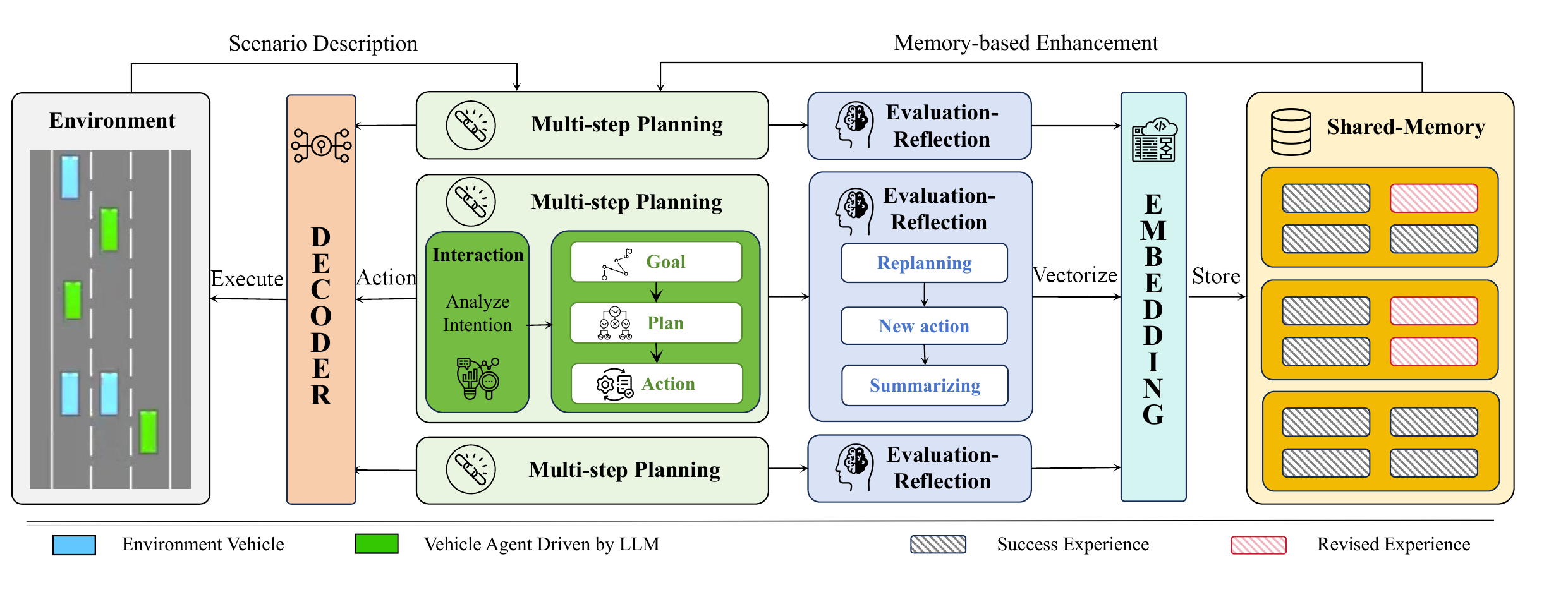}
    \caption{Knowledge-driven autonomous driving framework $\mathsf{KoMA}$ that incorporates multiple agents empowered by LLMs. $\mathsf{KoMA}$ consists of five core modules: the environment module, the multi-step planning module, the interaction module, the ranking-based reflection module, and the shared memory module. }
    \label{fig:1}
\end{figure*}

\subsection{LLM for Autonomous Driving}

The integration of LLMs into autonomous driving systems signified a transformative leap towards embedding human-like intelligence for enhanced decision-making and interaction capabilities.
Mao et al. emphasized the role of LLMs as cognitive agents in autonomous driving systems, highlighting their capacity to integrate human-like intelligence across various functions such as perception, prediction, and planning\cite{mao2023language}. This foundational approach underscored the potential of LLMs to mimic human cognitive processes.
Li et al. discussed the employment of LLMs as foundation models for autonomous driving, capitalizing on their rich repository of human driving experience and common sense. These models actively understood, interacted with, acquired knowledge from, and reasoned about driving scenarios\cite{li2023towards}.
Wen et al. introduced DiLu, a framework combining reasoning and reflection modules to facilitate decision-making based on common-sense knowledge while enabling continuous system evolution. DiLu's extensive experiments demonstrated its superior generalization ability over reinforcement learning-based methods, evidencing the capability of LLMs to accumulate experience and adaptively improve\cite{wendilu}.
Wang et al. proposed Co-Pilot, a universal framework that incorporates LLMs as a vehicle's ``Co-Pilot", adept at fulfilling specific driving tasks with human intentions in mind. This framework not only defined a workflow for human-vehicle interaction but also introduced a memory mechanism for task-related information organization and an expert-oriented black-box tuning to enhance performance without the need for fine-tuning the LLMs. The application of Co-Pilot in path tracking control and trajectory planning tasks showcased its versatility and effectiveness\cite{wang2023chatgpt}.
Shao et al. presented LMDrive, an innovative language-guided, end-to-end, closed-loop framework for autonomous driving. Uniquely integrating multi-modal sensor data with natural language instructions, LMDrive facilitated interaction with humans and navigation software in realistic settings\cite{shao2024lmdrive}.
    
The exploration of LLMs within autonomous driving systems unveiled a promising horizon where vehicles not only mimicked human driving capabilities but also engaged in complex decision-making and problem-solving tasks with an unprecedented level of intelligence and adaptability. The reviewed literature underscored the shift towards leveraging human-like intelligence and knowledge-driven methodologies, illustrating the potential of LLMs to redefine autonomous driving.However, current research on LLM-empowered autonomous driving agents mainly focused on simple scenarios lacking conflicts. We try to further explore and test the scope of knowledge driven capabilities in scenarios with conflicts, in order to fill the gap in this field.

\section{Methodology}
\subsection{Overview}
We introduce a knowledge-driven autonomous driving framework $\mathsf{KoMA}$ that incorporates multiple agents empowered by LLMs, comprising five integral modules: \textbf{\textit{Environment, Multi-agent Interaction, Multi-step Planning, Shared Memory, and Ranking-based Reflection}}. Within this framework, all agents operate on an equal and independent basis, as illustrated in Fig. \ref{fig:1}.

The environment module provides driving scenarios for the driving agent, which can be either a simulation environment or a real-world scenario. It is mainly responsible for providing a text description of the corresponding scene for each autonomous driving agent before making decisions.

The multi-agent interaction module further processes the textual information returned by the environment module, primarily enables the LLMs to analyze the behavior of other vehicles in the scenario like a human, infer their intentions, and support subsequent action decisions with relevant information.

Chain of Thought (CoT) is a technique used in LLMs that promotes complex reasoning and problem-solving capabilities. The core idea behind this method is to break down a complex problem into a series of smaller steps, known as intermediate reasoning steps. This enabless the model to incrementally construct a complete solution by addressing each step in a logical and sequential manner\cite{wei2022chain,lyu2023faithful,zhangmultimodal}. The multi-step planning module is an application of the CoT technique to guide LLM make the final action decision. The LLMs firstly analyze the goal according to the current scenario, then formulates the plan, and finally makes an action decision. This structured planning process enables the LLM agent to maintain a clear goal for its actions and more effectively pursue long-term goals. Based on the textual description of the current scene and the experiential playback of historical similar scenes, the LLM ultimately selects an action through continuous analysis and then returns the action decision to the environment module for execution.

The shared memory module utilizes a shared vector database to store the successful driving experiences of all agents. Before each agent makes a decision, the module retrieves analogous descriptions relevant to the agent's current situation and then provides these experiences to the agent, helping to formulate an informed action decision. This module allows each agent to train, accumulate experience, and interact with the environment, continuously enhancing decision-making effectiveness.

To ensure the quality of experiences in the shared memory module, this framework introduces a ranking-based reflection module that evaluates each driving decision after execution based on efficiency and safety. After a scenario has been concluded, the framework reviews the outcomes of those decisions, especially those with low scores or collisions. Only those experiences where decisions were corrected with high scores are retained. The procedure is outlined in Algorithm \ref{alg:alg1}.

\begin{algorithm}[!h]
\caption{Autonomous Driving with Multi-agent LLMs}\label{alg:alg1}
\begin{algorithmic}
\STATE 
\STATE {\textbf{Input}: Simulation scenario $sce$, simulation duration time $T$, current time $t$, decision interval time $\Delta$$t$, reflection agent $RA$, driving agent $DA$, driving agent list $DA\_list$, shared-memory $M$} 
\STATE {\textbf{Initialize}: \textit{$t$} = 0, score list $SL$ = [], decision list $DL$ = [], document $doc$ = [], $plan$ = None} 
\STATE {\textbf{while} \textit{$t$} $\textless$  \textit{$T$} \textbf{do} }
\STATE \hspace{0.5cm}{\textbf{for} $DA$ in $DA\_list$ :}
\STATE \hspace{0.5cm}\textcolor{cyan}{\textbackslash\textbackslash \hspace{0.25cm}Get the scenario description}
\STATE \hspace{1cm}$sce\_des$ = $sce$.describe(DA)
\STATE \hspace{0.5cm}\textcolor{cyan}{\textbackslash\textbackslash \hspace{0.25cm}Retrive similar experience from shared-memory}
\STATE \hspace{1cm}$few$-$shot$ = $M$.retriveMemory($sce\_des$)
\STATE \hspace{0.5cm}\textcolor{cyan}{\textbackslash\textbackslash \hspace{0.25cm}Driving agent making action decision}
\STATE \hspace{1cm}$action$, $plan$ = $DA$.reason($sce\_des$, $few$-$shot$, $plan$)\STATE \hspace{1cm}$DL$.append($action$)
\STATE \hspace{0.5cm}\textcolor{cyan}{\textbackslash\textbackslash \hspace{0.25cm}Execute the action decision in environment}
\STATE \hspace{1cm}$score$, $done$ = $env$.step($action$)
\STATE \hspace{1cm}$SL$.append($score$)
\STATE \hspace{1cm}$doc$.append([$DA$, $sce\_des$, $plan$, $action$, $score$])
\STATE \hspace{1cm}$t$ = $t$ + $\Delta$$t$
\STATE \hspace{0.5cm}\textcolor{cyan}{\textbackslash\textbackslash \hspace{0.25cm}Check whether the scene is finished}
\STATE \hspace{1cm}\textbf{if} $done$:
\STATE \hspace{1.5cm}$\textbf{break}$
\STATE \textcolor{cyan}{\textbackslash\textbackslash \hspace{0.25cm}Reflection agent correct the wrong action}
\STATE $corrected\_decision$ = $RA$.reflect($SL$, $DL$)
\STATE \textcolor{cyan}{\textbackslash\textbackslash \hspace{0.25cm}Update the memory}
\STATE $M$.update($corrected\_decision$)
\STATE {\textbf{Output}: Document $doc$, updated shared-memory $M$} 
\end{algorithmic}
\label{alg1}
\end{algorithm}

More details about the Multi-agent Interaction, Multi-step Planning, Shared Memory, and Ranking-based Reflection modules will be elaborated in the following sections.

\begin{figure*}[h]
    \centering
    \includegraphics[width=6in]{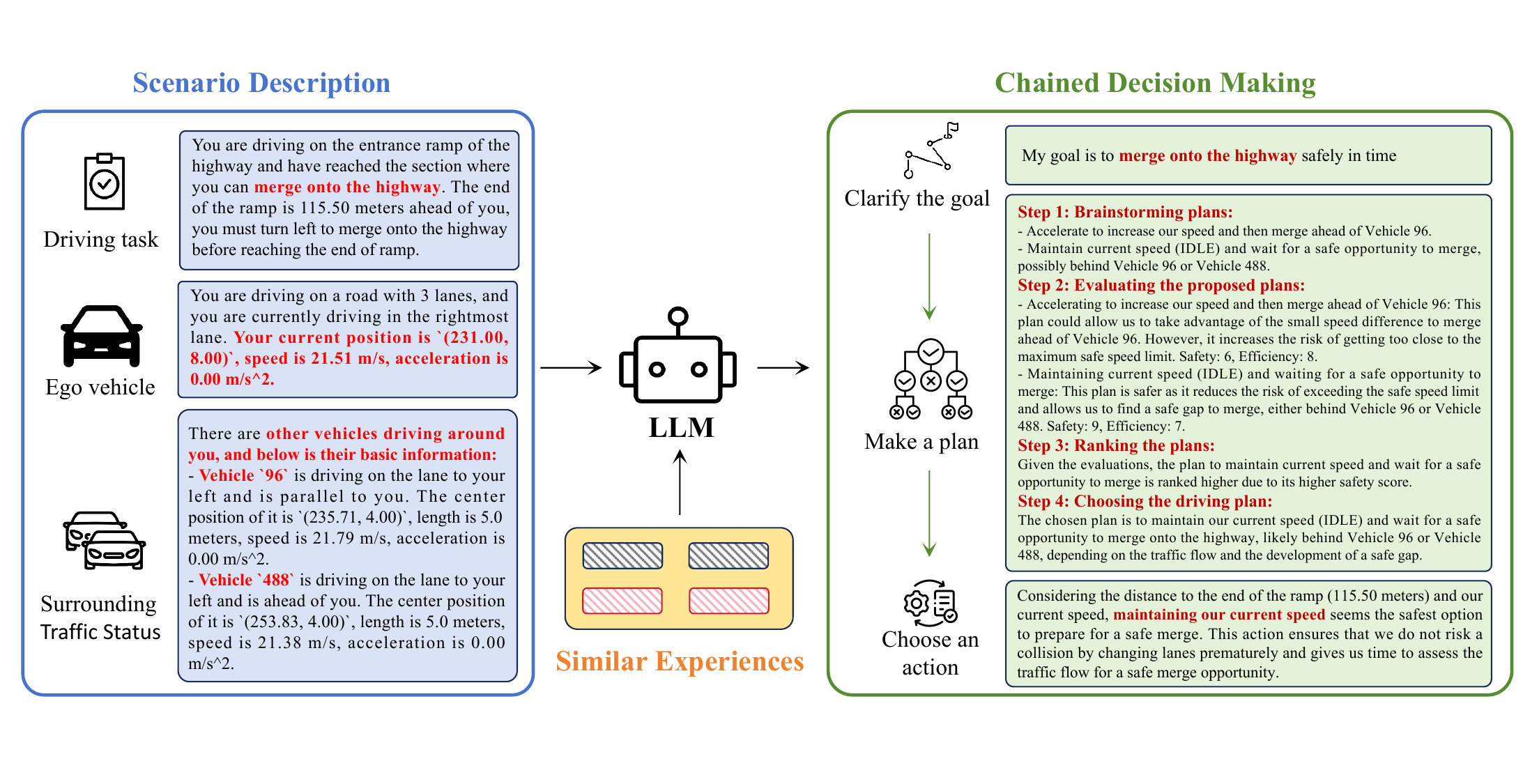}
    \caption{A case of multi-step planning module reasoning process. The multi-step planning module refers to a three-level multi-step reasoning of goal-plan-action, which analyzes and breaks down the scene target tasks step by step to ensure the consistency of the purpose of the decision-making before and after the action. Besides, it also refers to the four-step process of plan generation, plan evaluation, plan sorting, and plan selection when formulating a plan, to select the final plan that best fits the driving characteristics of the LLM, ensuring the feasibility and personalization of the plan.}
    \label{fig:2}
\end{figure*}

\subsection{Multi-agent Interaction Module}
In $\mathsf{KoMA}$, agents do not interact directly with each other; instead, agents infer their intentions just like real human drivers by analyzing the history and current state information of other vehicles. For example, when the LLM-driven agent senses a vehicle on its left suddenly accelerating, it further analyzes and reasons about this, guesses the vehicle's intention to overtake, and then analyzes these guesses along with its own plan to determine if adjustments to its short-term plan are needed. Ultimately, the agent aims to make rational action decisions. After receiving the text description from the environment module, the agent preliminarily processes the information using the multi-agent interaction module, analyzes and guesses the intentions of other vehicles before proceeding to the multi-step planning module with this information.

\subsection{Multi-step Planning}
The Multi-step Planning Module serves as the cornerstone for LLMs' reasoning within $\mathsf{KoMA}$. The process is illustrated in Fig \ref{fig:2}. Employing the Goal-Plan-Action methodology, this module merges inputs from the current scenario description with similar historical experiences to determine the best course of action, encompassing the following stages.

\textbf{Clarify goals based on the current scenario}:
Drivers have different goals to achieve in different scenarios. For example, the goal of a normal vehicle running on the highway is to drive efficiently and quickly under the premise of ensuring safety. The main goal of vehicles on the freeway ramp is to merge into the main road as soon as possible within a certain period of time, so the driver's goal is mainly related to the scenarios. Based on this, we divide the goals within the scenario into two categories: special scenario goals and general scenario goals. Special scenario goals refer to the tasks that need to be completed within a certain time in the scenario, such as ramp merging, intersection passage, etc. The general scenario goal is a long-term objective, such as maintaining safe and efficient driving. Each agent first sets its objectives based on the scenario description to inform planning and action decisions.

\textbf{Make a plan based on the goals}: The planning process involves the LLMs devising a strategy that connects identified goals with future actions, ensuring coherent actions over time. Initially, the LLMs checks for a pre-established plan. If no such plan exists, the LLMs formulates a new strategy to guide future actions. The planning process unfolds as follows:
\begin{enumerate}
\item The LLMs brainstorm all workable and distinct plans based on the current scenario and goal.
\item For each of the proposed plans, the LLMs will evaluate the potential of them, consider their pros and cons, implementation difficulty, potential challenges and then assign safety, efficiency score from 0 to 10 to each option based on these factors.
\item Based on the evaluations and scenarios, rank the plans.
\item A singular plan is selected to guide driving decisions.
\end{enumerate}
If there is a pre-existing plan, the LLM evaluates other vehicles' intentions from the multi-agent interaction module. If the plan is deemed infeasible, it craft a new plan; otherwise, it maintains the existing plan.

\textbf{Choose an action according to the plan}: It's time for LLMs making reasonable and safe action choices based on the current scene information and the plan. There are five actions in the action space:turn-left, IDLE (remain in the current lane with current speed), turn-right, acceleration, and deceleration. The LLMs select an action based on inference.

\begin{figure*}[h]
    \centering
    \includegraphics[width=6in]{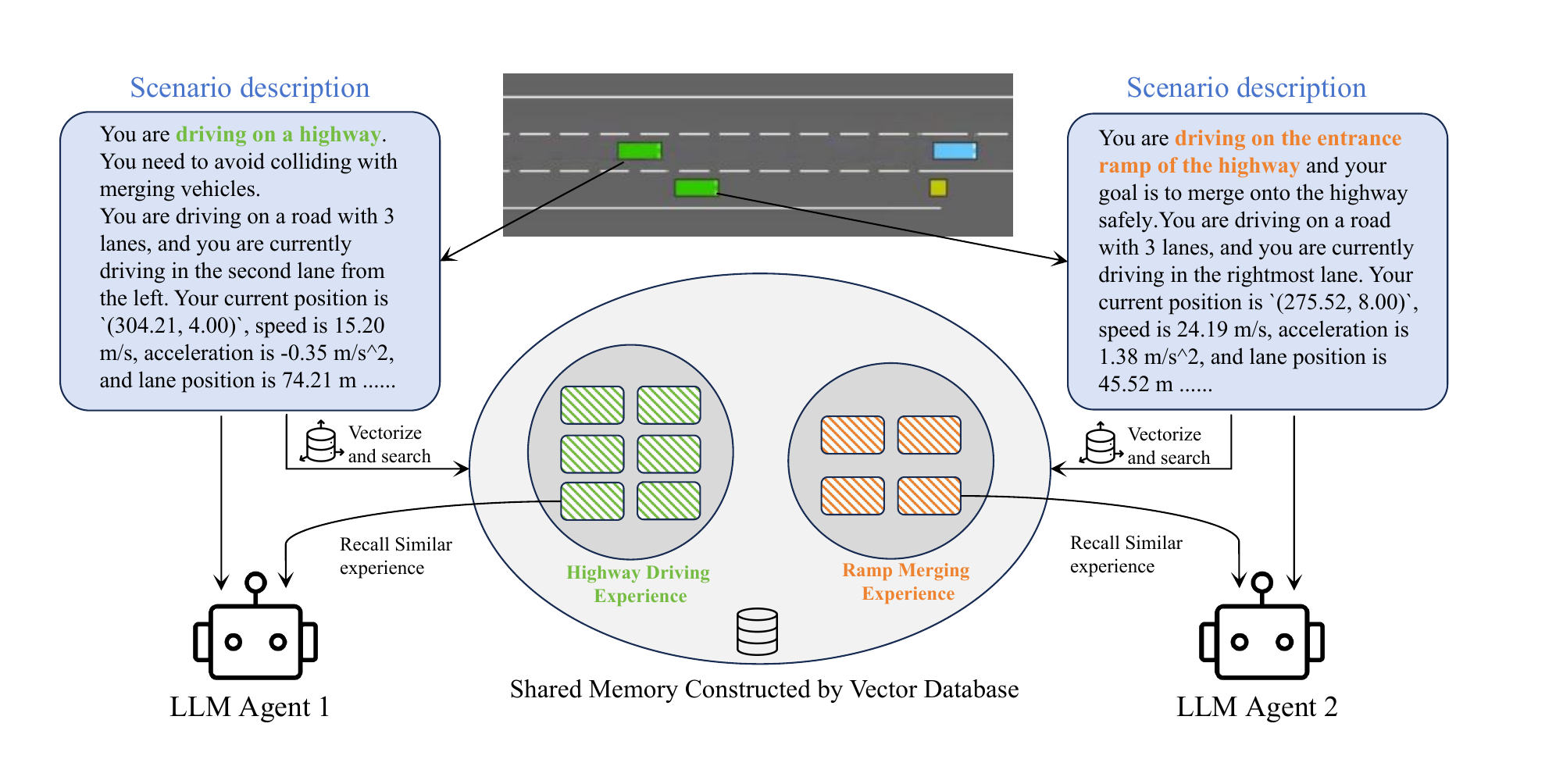}
    \caption{Different agents retrieve relevant experiences of their respective scenarios from shared memory modules. The shared memory module is a vector database that accumulates driving experience fragments from all agents. These fragments are vectorized and then stored in the same database. When making decisions, the agent retrieves similar driving experiences from analogous scenarios using vector search, thereby aiding in the decision-making process.}
    \label{fig:3}
\end{figure*}

\subsection{Shared Memory}
The shared memory module is an integral component of $\mathsf{KoMA}$, encapsulating a vector database designed to archive beneficial experiences. These archived experiences serve as exemplars to assist LLMs in the planning and decision-making processes. Each experience is segmented into four critical elements: the scenario description, the planning process, the final decision, and the evaluation score. The textual description of scenarios is converted into vectors, serving as keys for gauging similarity within the memory module. The concept of shared memory implies a unified memory module accessible to all agents empowered by LLM, fostering consistency of experience and performance. This approach mirrors the principle of parameter sharing observed in reinforcement learning.

The primary aim of shared memory is to quickly accumulate experiences in different scenarios simultaneously. Using the repository of past experiences, new agents can learn from the collective wisdom and insights of their predecessors. This shared memory module enables agents to perform tasks with an awareness of previously successful strategies and outcomes. Essentially, shared memory serves as a conduit for knowledge transfer, ensuring that the collective learning of agents is preserved and utilized to enhance future decision making.

\subsection{Ranking-based Reflection}
The reflection module, operational at the termination of each scenario, is designed to revise action decisions. These rectified decisions, along with successful experiences, are then integrated into the memory module. When the reflection module updates the reflection results to the shared memory module, it can be considered that the agent has completed one round of training. To differentiate the efficacy of each action decision, the system incorporates safety indicators and efficiency scores, assessing the vehicle's state post-execution of LLM-generated instructions. In the following, we delineate the criteria for these evaluations.

\begin{figure*}[hb]
    \centering
    \includegraphics[width=6in]{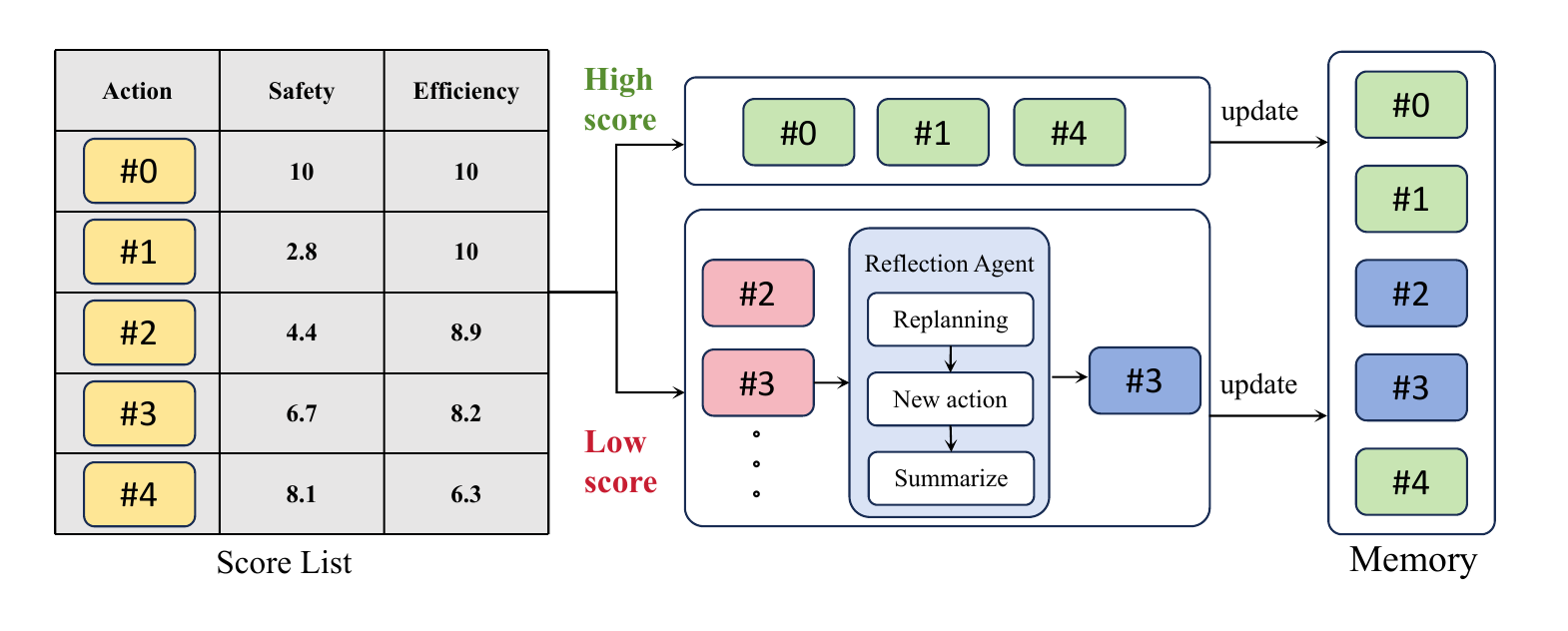}
    \caption{The ranking-based reflection module evaluates decisions, identifies those with low scores, and corrects them. It then updates the shared memory module with these refined decisions, along with the high-scoring experiences.}
    \label{fig:4}
\end{figure*}

\textbf{Safety evaluation criteria}: Safety is quantified on a scale from 0 to 10, using Time To Collision (TTC) to measure vehicular safety post-action. The TTC above 3 seconds indicates optimal safety, earning the highest score of 10. Below 1.5 seconds, TTC indicates a critical safety risk, scoring 0\cite{TTC1, TTC2, TTC3}. For TTC values between 1.5 and 3 seconds, the safety score linearly scales from 0 to 10.
\begin{equation*}
{Safety_{score}(t)} = \begin{cases}
\frac{20(TTC(t)-1.5)}{3},&{\text{if}}\ 1.5s<TTC(t)<3s \\ 
10,&{\text{else if}}\ TTC(t)>3s \\
{0,}&{\text{otherwise.}} 
\end{cases}
\end{equation*}

\textbf{Efficiency evaluation criteria}: Efficiency is similarly rated on a scale from 0 to 10 and uses speed as the efficiency metric. This metric considers the impact of surrounding vehicle speeds on the agent's vehicle, measuring efficiency by the difference between the agent's speed and the average speed of surrounding vehicles. An agent vehicle that matches or exceeds the average speed of surrounding vehicles receives a full efficiency score of 10. If the agent's speed is below the average, the efficiency score is the ratio of the agent's speed to the average speed, scaled to 10.
\begin{equation*}
{Efficiency_{score}(t)} = \begin{cases}
10,&{\text{if}}\ V_L{}_L{}_M(t)>V_A{}_v{}_g(t) \\ 
{0,}&{\text{otherwise.}} 
\end{cases}
\end{equation*}

Upon the conclusion of a scenario, the LLM undertakes an analysis of the scoring list to pinpoint actions that deviated from expected outcomes and requires reflection. This process mandates the correction of previously unsafe decisions, facilitating a cycle of continuous enhancement in the agent's capabilities. The utilization of interpretable chain-of-thought responses significantly aids in uncovering the root causes of potentially hazardous scenarios.

In an effort to foster autonomous learning from past mistakes, our methodology employs the detailed description of the driving scenario in which the erroneous decision was made, alongside the corresponding reasoning output, as inputs for the LLM. This approach prompts the LLM to elucidate the underlying reasons for the flawed decisions, thereby guiding the agent towards more accurate and safer future decisions. Additionally, the LLM is tasked with devising strategies to mitigate the recurrence of similar errors, enhancing its decision-making framework.This reflective process incorporates refined reasoning and revised decisions learned from error correction into memory modules. This ensures the preservation of enhanced knowledge and underlies the agent's ability to learn adaptively in a variety of driving environments.

\section{Experiments}
\subsection{Experimental Settings}

\textbf{Simulation Environment}: We leverage the ``highway-env" as the simulation platform, which furnishes a realistic multi-vehicle interactive environment\cite{leurent2018environment}. This environment permits the customization of various parameters including vehicle positioning, velocity, count, and lane specifics, offering a versatile setting for conducting our studies. The focal point of our scenario selection is the on-ramp merging challenge within a highway context. This particular scenario underscores the exigency of accomplishing merging objectives within a constrained timeframe, thereby serving as an apt representation of the agent's decision-making acumen.

\textbf{Large Language Model}: GPT-4\cite{achiam2023gpt} was mainly used in experiments to verify the validity of the $\mathsf{KoMA}$ framework. This model plays a pivotal role in both the Multi-step Planning and Rank-based Reflection modules of our architecture, showcasing its versatility and advanced understanding capabilities. For the Memory module, integral to our framework, we incorporate ``Chroma" an open-source embedded vector database. This choice is instrumental in facilitating the conversion of scene descriptions into vectors, for which we employ OpenAI's ``text-embeddings-ada-002" model. This comprehensive setup not only exemplifies the integration of cutting-edge AI technologies but also highlights our innovative approach towards enhancing autonomous driving systems through nuanced understanding and reflection. TABLE \ref{tab:table1} shows the parameters and components related to $\mathsf{KoMA}$.
 
\begin{table}[h]
\caption{the parameters and components related to $\mathsf{KoMA}$\label{tab:table1}}
\centering
\begin{tabular}{|c||c|}
\hline
\textbf{Parameter} & \textbf{Value}\\
\hline
Simulation environment & highway-env\cite{leurent2018environment}\\
\hline
Large language model & GPT-4\cite{achiam2023gpt}\\
\hline
Text embedding model & text-embeddings-ada-002\\
\hline
Vector database & Chroma\\
\hline
\end{tabular}
\end{table}

\begin{figure}[h]
    \centering
    \subfigure[An example of initialization scenario]{
    \includegraphics[width=0.9\linewidth]{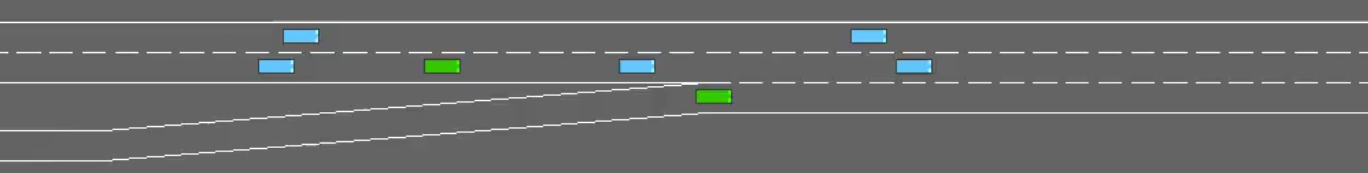}}
    \subfigure[A failure scenario where a collision occurs on a ramp merge]{
    \includegraphics[width=0.9\linewidth]{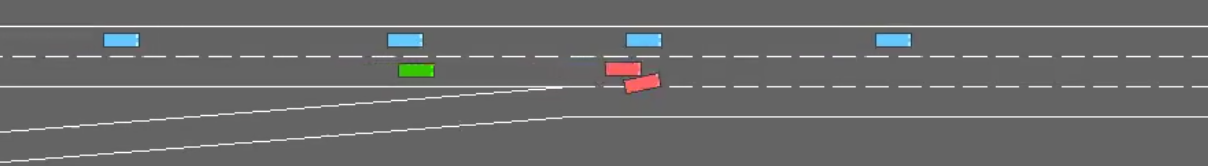}}
    \subfigure[A failure scenario where the ramp merge is not completed in time]{
    \includegraphics[width=0.9\linewidth]{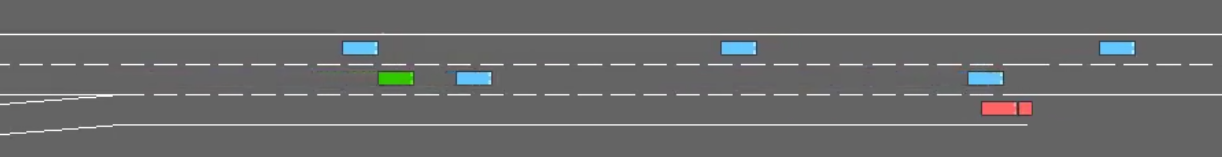}}
    \subfigure[A successful scenario for safely completing ramp merges]{
    \includegraphics[width=0.9\linewidth]{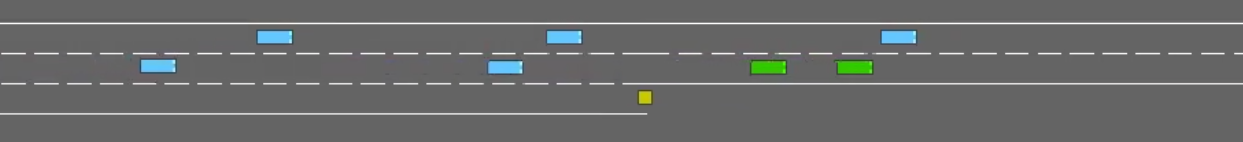}}
    \caption{Examples of different phases in the testing scenarios }
    \label{scenario}
\end{figure}

\textbf{Testing Scenario}: The testing scenario locates at an on-ramp entrance on the highway with two main lanes. The scenario incorporates two LLM-driven agents, one on the main road and the other on the on-ramp, as shown in Fig \ref{scenario}. 
When the agent reaches the endpoint or collides with other vehicles, it is considered as the end of the scenario. The reflection module will then update the shared-memory module. When the update is completed, the agent completes a round of training in that scene. After every 20 rounds of training, 20 randomly generated scenarios are selected and fixed for testing the trained agents in $\mathsf{KoMA}$. 
\begin{table}[h]
\caption{the parameters of the initialization scenarios\label{tab:table2}}
\centering
\begin{tabular}{|c||c|}
\hline
\textbf{Parameter} & \textbf{Value}\\
\hline
The lane count of the highway main road & 2\\
\hline
The lane count of the on-ramp merging road & 1\\
\hline
The length of on-ramp merging road & 120 m\\
\hline
The initialization speed of vehicles & range from 20 to 25 m/s\\
\hline
The spacing of the spawn points & 40 m\\
\hline
The initial location noise & range from -10 to 10 m\\
\hline
The count of LLM controlled vehicles & 2\\
\hline
The count of intelligent driver model vehicles & 5\\
\hline
Policy frequency & 2 Hz\\
\hline
\end{tabular}
\end{table}

\textbf{Baseline MARL-based model}: The baseline MARL method is ``deep multi-agent reinforcement learning for highway on-ramp merging in mixed traffic"\cite{chen2023deep}, which develop an efficient and scalable MARL framework that can be used in dynamic traffic where the communication topology could be time-varying. We have adjusted the MARL parameter Policy frequency from 5Hz to 2Hz to maintain consistency with the Policy frequency during training in $\mathsf{KoMA}$. Concurrently, we have adapted the MARL training scenarios to focus on the two-lane highway on-ramp merging context and have initiated the training process accordingly.

\subsection{Performance of the proposed framework}
In this section, we conducted comparative experiments on $\mathsf{KoMA}$ and deep multi-agent reinforcement learning (MARL). To maintain consistency with the test scenarios of the MARL methods, we conducted comparative tests in scenarios where both the main road and the highway ramp are configured as single lanes.\par

\begin{figure}[bh!]
    \centering
    \includegraphics[width=0.9\linewidth]{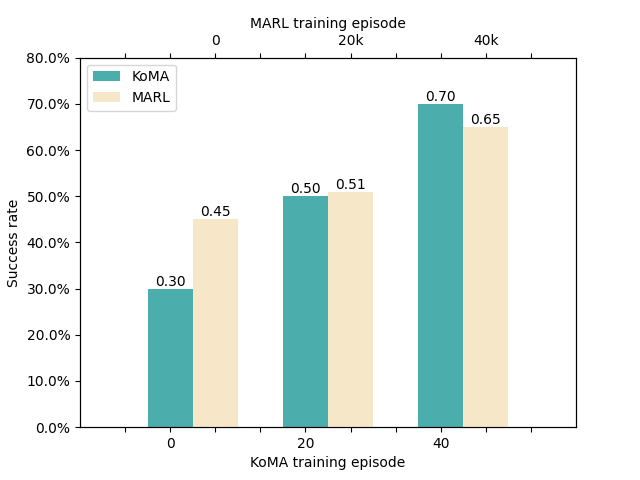}
    \caption{The performance of $\mathsf{KoMA}$ after 0, 20, and 40 training episodes is compared with the performance of MARL after 0, 20,000, and 40,000 training episodes, respectively.}
    \label{train3}
\end{figure}

\textbf{Performance}: In the untrained state, the success rate of the $\mathsf{KoMA}$ framework is 30\%, while the success rate of MARL reaches 45\%. The success rate of MARL is slightly higher than that of the $\mathsf{KoMA}$ framework. This is because the selected GPT4 model is a general LLM that has not been extensively trained with corpora in the driving domain, resulting in a less clear understanding of driving tasks. However, after 20 rounds of training and accumulating driving experience, the $\mathsf{KoMA}$ framework has achieved a success rate of 50\%, which is essentially on par with the 51\% success rate achieved by the MARL algorithm after 20,000 rounds of training. When the $\mathsf{KoMA}$ framework has undergone 40 rounds of training, it has already achieved a success rate of 70\%, successfully surpassing the 65\% success rate achieved by the MARL algorithm after 40,000 rounds of training. This experiment not only demonstrates the effectiveness of the $\mathsf{KoMA}$ framework in closed-loop learning within scenarios but also highlights that, compared to traditional data-driven methods, knowledge-driven agents possess higher learning efficiency and better training outcomes.
\subsection{The validation of the framework}

\subsubsection{Shared Memory}

\begin{figure}[h]
    \centering
    \includegraphics[width=0.9\linewidth]{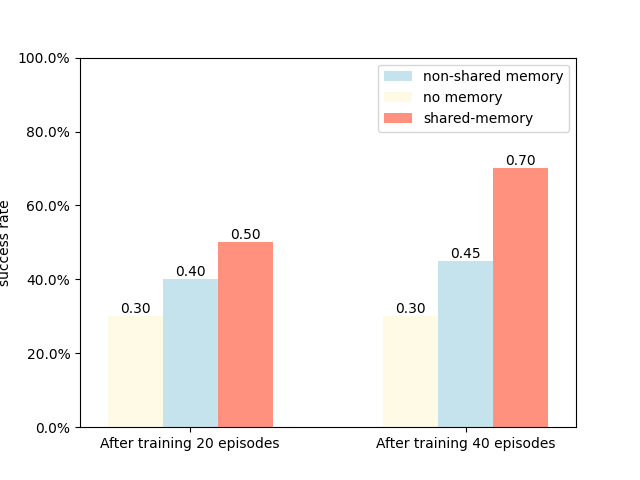}    
    \caption{Experimentalresults tested in the initial scenario with different memory modules after 20 and 40 rounds of training.}
    \label{no_shared}
\end{figure}

In this section, an ablation study is conducted on the Shared Memory module. We have set up three different types of memory modules for comparative testing to verify the superiority of the shared memory module. The first type is the no-memory module, which means that no vector database is set up to accumulate and store historical experiences. The second type is the non-shared memory module, where each agent has its own vector database, and the accumulated experience fragments are not shared with other agents, only used to assist in their own action decisions. The third type is the shared memory module, where each agent shares a common vector database, and the experiences accumulated by an agent can be called upon by other agents to assist in their action decisions, thus achieving sharing. The testing results are shown in Fig. \ref{no_shared}.

The non-shared memory module exhibits poorer performance during training; after 40 rounds of training, the success rate is only 45\%, which is worse than the performance of the shared memory module after just 20 rounds of training. This is partially because sharing a common memory module leads to a richer and more abundant vector database, allowing for the simultaneous accumulation of training effects from multiple agents. After the vehicles driving on ramp merge onto the highway, they still need to travel on the main road for some time, which requires the agent to have a certain amount of highway driving experience. However, most of the experience of agents on the ramp is confined to the ramp scenario, with less experience driving on the main road, which requires more training rounds. Therefore, the shared memory module not only effectively enhances the training effectiveness of multiple agents, improving collective intelligence, but also ensures that all agents are at the same level of intelligence, preventing the group's performance from being adversely affected by poorly trained agents.

\subsubsection{Multi-step Planning}
In this section, we will compare the training results of agents with and without the multi-step planning module to verify the effectiveness of $\mathsf{KoMA}$. The agents under these two frameworks are both trained for 40 rounds. The testing results are displayed via a boxplot in Table \ref{tab:table3}. It shows the success rates of agents at the different training stages, including agents without training, agents trained with 20 rounds and agents trained with 40 rounds. 

\begin{table}[h]
\caption{The success rate of experimental results in the initial scenario was evaluated using different frameworks, with testing conducted after 0, 20, and 40 rounds of training.\label{tab:table3}}
\centering
\begin{tabular}{|c||c||c|}
\hline
\textbf{Training episodes} & \textbf{DiLu} & \textbf{KoMA}\\
\hline
0 & \textbf{40\%} & 30\%\\
\hline
20 & 50\% & \textbf{50\%}\\
\hline
40 & 50\% & \textbf{70\%}\\
\hline
\end{tabular}
\end{table}

\begin{figure}[h]
    \centering
    \includegraphics[width=0.9\linewidth]{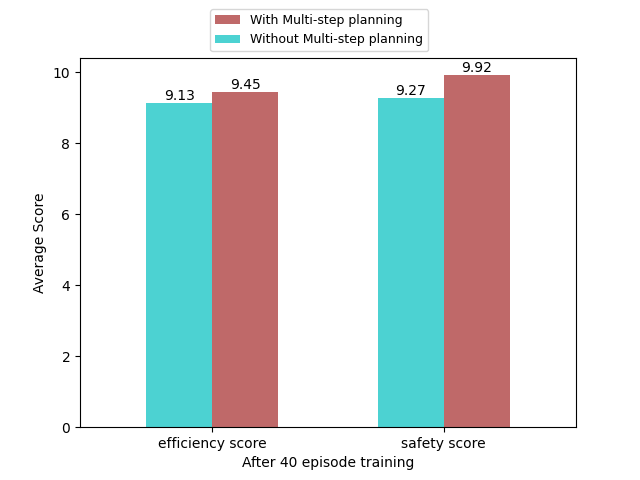}
    \caption{The average efficiency score and safety score of successful testing scenarios after 40 rounds of training.}
    \label{train2}
\end{figure}
We found that agents with multi-step planning modules performed worse without training, with a success rate of only 30\%. However, during 20 rounds of training, the effect was on par with that of agents without multi-step planning modules, achieving a success rate of 50\%. After 40 rounds of training, the agent with multi-step planning modules achieved a success rate of 70\%. This indicates that agents with multi-step planning modules have better training effects and faster convergence in ramp merging scenarios. The Fig. \ref{train2}
shows the average score of successful testing scenarios after 40 rounds of training. Agents with multi-step planning modules have better efficiency and safety scores, indicating that they can complete scenario testing more safely and efficiently. This may be because the three-layer progressive reasoning thinking chain of goal-plan-action can better assist the LLM in making action decisions with consistent goals when completing urgent tasks in complex and time-varying scenarios.When there is a clear short-term goal, action decisions become more purposeful, avoiding meaningless action decisions of repeated acceleration and deceleration, which improves the efficiency of the agent. Meanwhile, the inspection part of the plan ensures that the agent can timely reformulate the plan when the old plan is no longer safe and feasible in time-varying scenarios, avoiding the continued execution of high-risk old plans and improving the security of the agent.


\subsection{Generalization under different Scenarios}
Data-driven agents often tend to overfit to training scenarios, leading to poor generalization capabilities and limited applicability\cite{li2023towards, ying2019overview, hawkins2004problem}. Therefore, in this section, we conduct a series of tests on the generalization capabilities of knowledge-driven agents within the framework to examine whether LLMs can learn knowledge from experiential fragments and apply it to different scenarios.

\subsubsection{Altering the number of lanes on the main roadway}
\begin{figure}[h]
    \centering
    \includegraphics[width=0.9\linewidth]{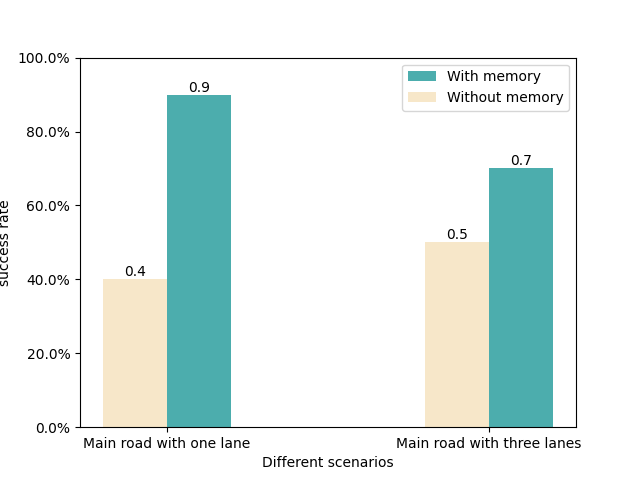}
    \caption{The experimental results for different generalization scenarios are evaluated with and without the memory, which has been trained on a main roadway with two lanes.}
    \label{train3}
\end{figure}

Initially, we conducted tests with minor variations in scenario complexity by changing the number of lanes on the merging main roadway. Despite these minor changes, these scenarios still pose challenges for data-driven agents in generalizing. We trained the LLM agents on scenarios with two-lane main roadways and single-lane ramps. After training, we obtained an experience-rich vector database, known as the memory module. We then tested this module in scenarios with one more or one less lane on the main roadway, at the same density. As the memory module lacks content for these new scenarios, this setup tests the LLM agents' ability to generalize. The results are shown in Fig. \ref{train3}. We found that the shared-memory trained in the initial scenario still affects and yields good results when generalizing to scenarios with varying main road lanes, indicating superior generalization ability in knowledge-driven agents compared to traditional data-driven ones.

\subsubsection{Generalizing to roundabout scenarios}
Continuing with the use of the same memory module as in the previous section, we shifted the scenario from ramp merging to roundabouts. Our aim was to test whether the LLM agents could learn from previous scenarios and enhance their decision-making abilities by generalizing to completely different scenarios. Additionally, this shift aimed to demonstrate the value of a shared memory module in enhancing the generalization capabilities of multiple agents.

\begin{figure}[ht]
    \centering
    \subfigure[The first phase of the roundabout driving scenario: Gradually approach the roundabout and observe the traffic conditions within it, looking for the right moment to enter the roundabout.]{
    \includegraphics[width=0.9\linewidth]{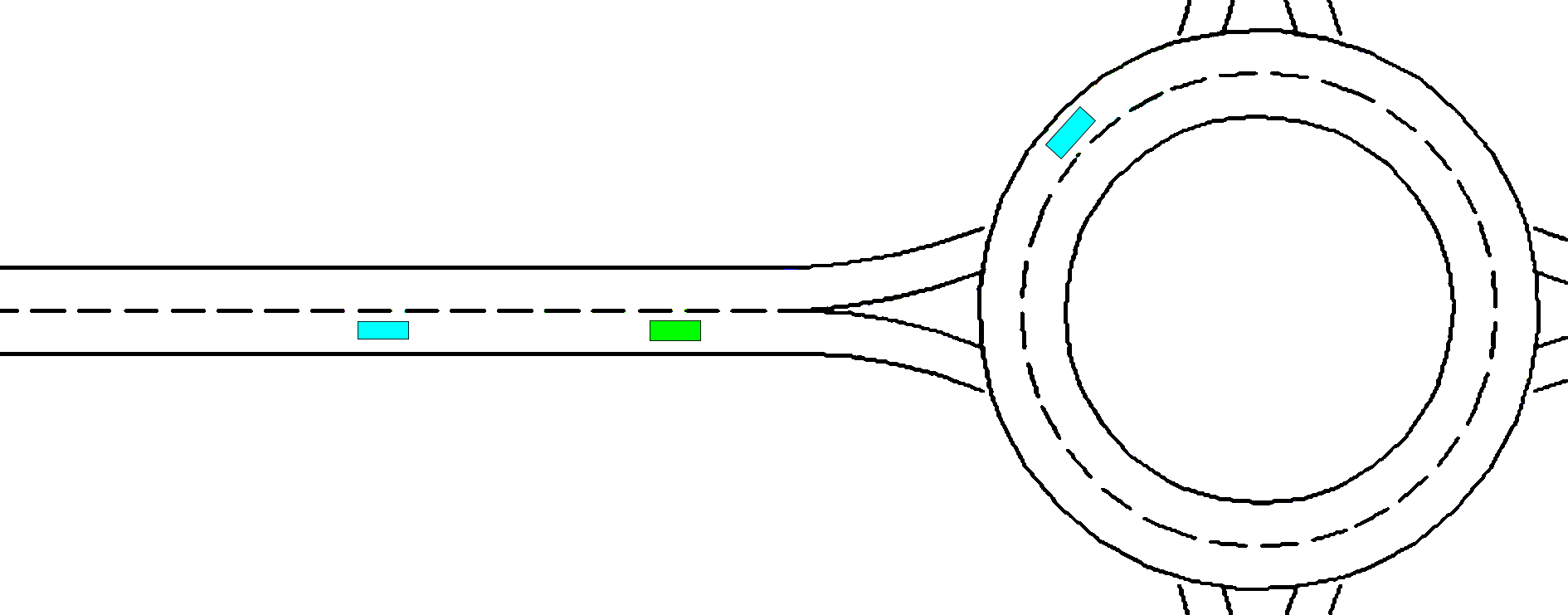}}
    \subfigure[The second phase of the roundabout driving scenario: The vehicle successfully enters the roundabout and drives within it. The vehicle needs to identify its exit within the roundabout and merge out in a timely manner at the designated exit.]{
    \includegraphics[width=0.9\linewidth]{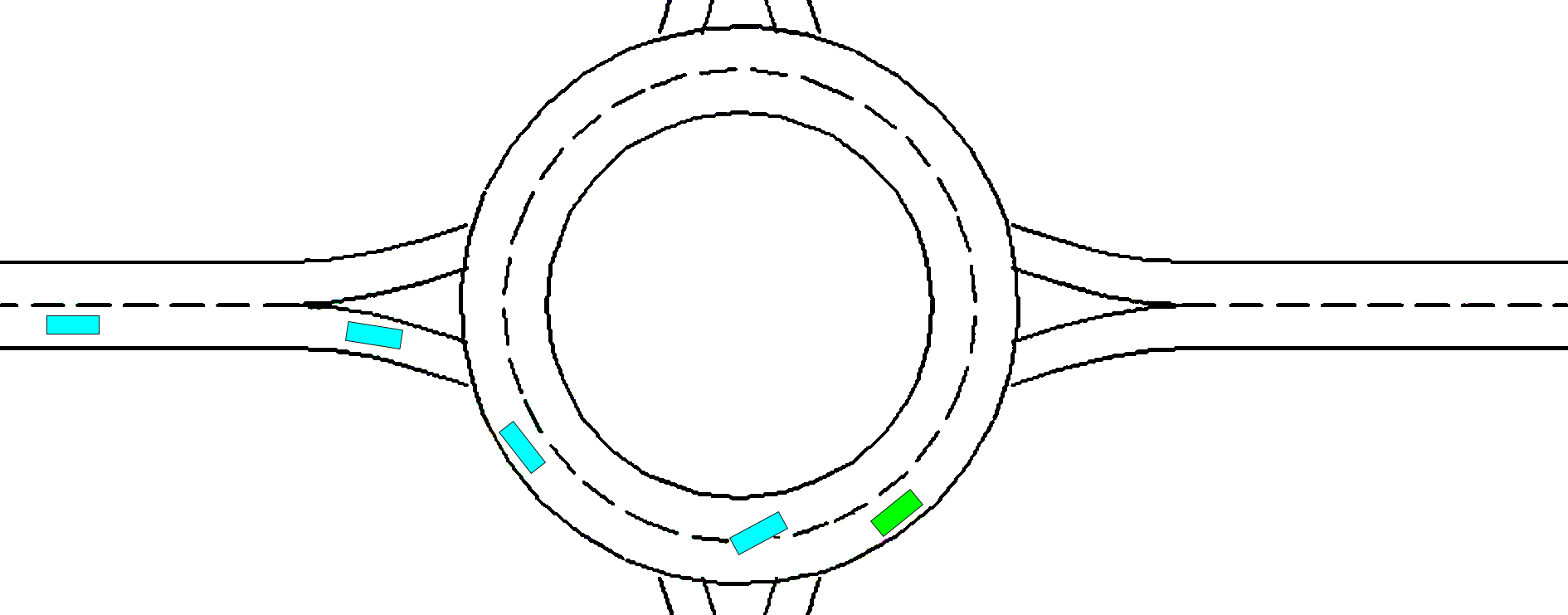}}
    \subfigure[The third phase of the roundabout driving scenario: The vehicle successfully merges out from the exit and leaves the roundabout, completing the roundabout scenario.]{
    \includegraphics[width=0.9\linewidth]{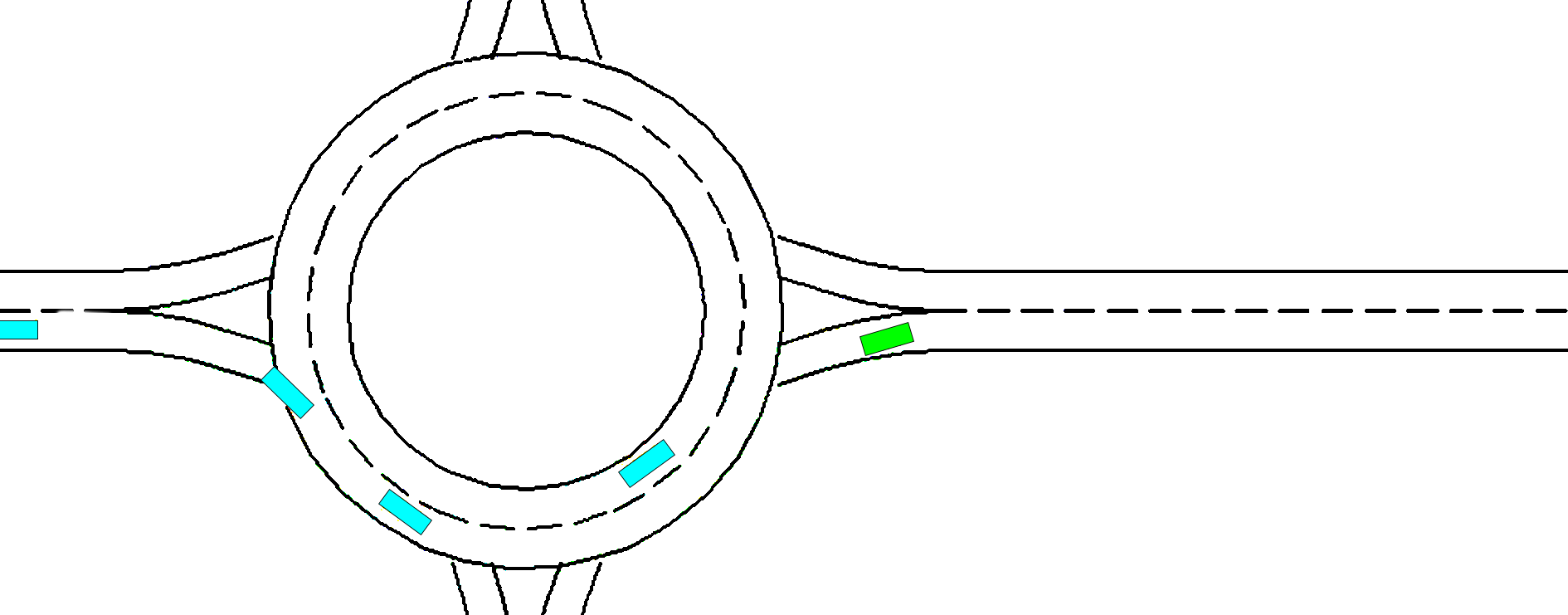}}
    \caption{Illustrations of the three stages of the roundabout scenario.}
    \label{roundabout_stage}
\end{figure}

\begin{figure}[ht]
    \centering
    \includegraphics[width=0.9\linewidth]{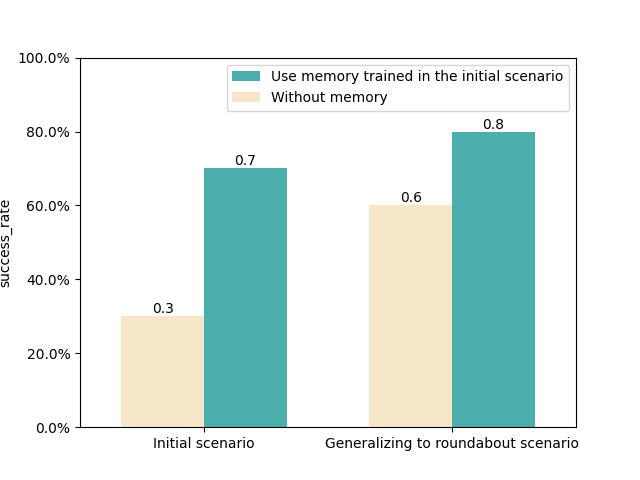}
    \caption{The generalization ability of the memory, trained for 40 episodes in the initial scenario, is specifically tested in the roundabout scenario.}
    \label{roundabout_v2}
\end{figure}

The roundabout scenario setup is depicted in Fig. \ref{roundabout_stage}. There is a single lane for each of the four entry and exit roads of the roundabout. Within the roundabout itself, there is a two-lane circular road. Roundabout scenarios can be mainly divided into three phases: 1. Roundabout entry phase: Merging into the roundabout from the entry road. 2. Roundabout internal driving phase: Driving within the roundabout until approaching the target exit road. 3. Roundabout exit phase: Exiting from the roundabout to the exit road and leaving the roundabout. There are four vehicles modeled as intelligent driver model (IDM) on the western entry road of the roundabout, with their destination being the eastern exit road. The agent vehicle controlled by the LLM is on the southern entry road, followed by an IDM environmental vehicle. There are a total of six vehicles in the entire roundabout scenario, and their actual generation positions still follow the form of fixed coordinate points plus random fluctuations to reflect randomness. In this scenario, the agent vehicle controlled by the LLM needs to safely merge into the roundabout from the southern entry and smoothly exit from the eastern side of the roundabout, which is considered a successful completion of the scenario. If a collision occurs or the vehicle fails to exit from the eastern side of the roundabout in time, it is considered a failure of the scenario. The result are shown in Fig. \ref{roundabout_v2}

In the initial ramp merging scenario on a two-lane main road, the success rate was 30\% without prior training, and it reached a 70\% success rate after 40 rounds of training. An agent without any memory repository achieved a 60\% success rate in the roundabout scenario. When the memory repository from 40 rounds of training in the initial ramp merging scenario was applied to the agent in the roundabout scenario, the success rate increased to 80\%, demonstrating the effectiveness of knowledge-driven generalization capabilities of the agent. The reason for the roundabout scenario still having a 60\% success rate without experience is mainly twofold: firstly, because there is only one LLM controlling the agent in the scenario, which reduces the overall collision risk of the LLM agent; secondly, because the roundabout scenario is simpler, with only longitudinal acceleration and deceleration control before merging, without the need to consider lane-changing behavior.

\subsection{Performance under different LLMs}
In this section, we will employ various LLMs to verify the efficacy of $\mathsf{KoMA}$ framework. By comparing their performance during training, we aim to ascertain which model is better suited for our $\mathsf{KoMA}$ framework.

\begin{table}[h]
\caption{The success rate of experimental results with different LLMs after 0, 20, 40 rounds of training in the KoMA framework\label{tab:table4}}
\centering
\begin{tabular}{|c||c||c||c||c||c||c|}
\hline
\textbf{episodes} & \textbf{GPT3.5} & \textbf{GPT4} & \textbf{Llama3-8B} & \textbf{Llama2-7B} & \textbf{Qwen2-7B}\\
\hline
0 & 20\% & \textbf{30\%} & 25\% & 20\% & 25\%\\
\hline
20 & 30\% & \textbf{50\%} & 35\% & 25\% & 30\%\\
\hline
40 & 35\% & \textbf{70\%} & 40\% & 35\% & 40\%\\
\hline
\end{tabular}
\end{table}

Our research findings demonstrate that a range of Large Language Models (LLMs), including GPT3.5\cite{brown2020language}, GPT4\cite{achiam2023gpt}, LLaMA2-7B\cite{touvron2023llama}, LLaMA3-8B\cite{Llama3}, and QWEN2-7B\cite{bai2023qwen}, can be effectively trained within the $\mathsf{KoMA}$ framework. This training significantly improves their decision-making capabilities, thereby validating the framework's universal applicability and effectiveness across different models. GPT-4 performs best within the $\mathsf{KoMA}$ framework, with better training outcomes and faster convergence speed. GPT-3.5 underperforms within $\mathsf{KoMA}$, particularly in handling lengthy scenario texts. The model's difficulty in accurately capturing critical details, such as vehicle spacing and speed variations, often leads to the exclusion of essential information during decision-making processes. This oversight can result in collisions involving the agent vehicles, highlighting the need for enhanced model training and attention to detail recognition.

\section{Conclusion}

The advent of LLMs as autonomous agents has marked a significant shift in the approach to knowledge-driven problem-solving, particularly within the autonomous driving sector. This study presents the $\mathsf{KoMA}$ framework, a comprehensive system designed to transcend the limitations of current single-agent models by facilitating multi-agent collaboration, shared knowledge, and cognitive synergy. The $\mathsf{KoMA}$ framework's multi-faceted approach, integrating interaction, planning, memory, and reflection modules, has proven instrumental in enhancing decision-making in complex driving scenarios. Our empirical evaluations have confirmed the $\mathsf{KoMA}$ framework's superiority over traditional methods, showcasing its robustness in unpredictable environments and its ability to generalize across a wide array of driving situations without the need for extensive retraining. The framework's success lies in its human-like cognition, where LLM agents can discern and respond to the intentions of surrounding vehicles, and its layered planning ensures strategic consistency in action decisions.

The $\mathsf{KoMA}$ framework is a preliminary test of the potential of LLMs in revolutionizing autonomous systems. It opens a new frontier for research, suggesting that the future of autonomous driving lies in the seamless integration and collaboration of multiple agents. As we continue to refine and expand upon this framework, we envision a future where autonomous driving systems are not only highly adaptable and safe but also capable of learning and improving continuously from shared experiences and insights.

\section*{Acknowledgments}
This work was also partially supported by the National Natural Science Foundation of China (project number:
52202378), Beijing Natural Science Foundation (project number: L243008), the Open Research Project Program of the State Key Laboratory of Internet of Things for Smart City (project number: SKL-IoTSC(UM)-2021-2023/ORP/GA08/2022), and the Ministry of Transport of PRC Key Laboratory of Transport Industry of Comprehensive Transportation Theory (Grant No. MTF2023002).

\bibliographystyle{IEEEtran}
\bibliography{reference}

\newpage

 
\vspace{11pt}


\vspace{11pt}


\vfill

\end{document}